\documentclass[conference]{IEEEtran}
\IEEEoverridecommandlockouts

\usepackage{tabto}
\usepackage{algpseudocode}
\usepackage{algorithm}
\usepackage{paralist}
\usepackage{cite}
\usepackage{amsmath,amssymb,amsfonts}
\usepackage{graphicx}
\usepackage{textcomp}
\usepackage{xcolor}
\usepackage{hyperref}
\usepackage{float}
\usepackage{subcaption}
\usepackage{enumitem}
\usepackage{amsmath}
\usepackage{placeins}
\def\BibTeX{{\rm B\kern-.05em{\sc i\kern-.025em b}\kern-.08em
    T\kern-.1667em\lower.7ex\hbox{E}\kern-.125emX}}

\usepackage{hyperref}

\hypersetup{
    colorlinks=true,
    linkcolor=blue,
    filecolor=magenta,
    urlcolor=cyan,
    pdftitle={ },
    pdfpagemode=,
    }

\newcommand{\nix}[1]{}

\begin{document}

\title{{Early Diagnosis of Acute Lymphoblastic Leukemia Using YOLOv8 and YOLOv11 Deep Learning Models}

}

\author{Alaa Awad$^\S$ ~~ and ~~  Salah A. Aly$^\dag$$^\ddag$ \\
$^\S$CSE Dept., E-Japan University of Science and Tech., Alexandria, Egypt\\
$^\dag$Faculty of Computing and Data Science, Badya University, Giza, Egypt\\
$^\ddag$Computer Science Sect., Faculty of Science, Fayoum University, Fayoum, Egypt
}

\maketitle
\begin{abstract}
Leukemia, a severe form of blood cancer, claims thousands of lives each year. This study focuses on the detection of Acute Lymphoblastic Leukemia (ALL) using advanced image processing and deep learning techniques. By leveraging recent advancements in artificial intelligence, the research evaluates the reliability of these methods in practical, real-world scenarios. Specifically, it examines the performance of state-of-the-art YOLO models, including YOLOv8 and YOLOv11, to distinguish between malignant and benign white blood cells and accurately identify different stages of ALL, including early stages. Moreover, the models demonstrate the ability to detect hematogones, which are frequently misclassified as ALL. With accuracy rates reaching 98.8\%, this study highlights the potential of these algorithms to provide robust and precise leukemia detection across diverse datasets and conditions.
\end{abstract}


\section{Introduction}
We live in a technology-driven era where computer science plays a crucial role in replicating human intelligence to enable faster and more accurate decision-making. Leveraging this, many researchers have explored leukemia detection using Artificial Intelligence (AI) through deep learning methods like MobileNetV2~\cite{Sandler2019}, attention mechanisms~\cite{Vaswani2023}, and YOLO~\cite{Redmon2016}. Various datasets, such as the ALL image dataset~\cite{Ghaderzadeh2021} and the C-NMC 2019 dataset~\cite{Mourya2019}, have been utilized to advance these efforts.

Leukemia, commonly referred to as blood cancer, is one of several cancers that start in the bone marrow and blood. The fast and aberrant synthesis of white blood cells is the cause of this problem. Depending on how quickly the disease progresses, leukemia can be classified as either chronic or acute. It can also be classified as lymphocytic or myelogenous depending on the type of cells it changes into after growth, which can be limited to white lymphocyte blood cells or myelogenous, which can then be of one of three types: red blood cells, white blood cells, or platelets~\cite{who2023}.

Despite these advancements, most existing studies rely on single-cell datasets for training AI models. However, real-world scenarios often involve multi-cell images, presenting a challenge for models to maintain high accuracy. This paper addresses this limitation by training models on multi-cell samples to improve their practical applicability.

To achieve this, we employed image processing techniques, including segmentation, to prepare the dataset. Furthermore, transfer learning and fine-tuning were applied to models like YOLOv11~\cite{YOLOv11_2024} and YOLOv8~\cite{Varghese2024}, resulting in accuracies exceeding 98

\begin{figure}[h]
\centerline{\includegraphics[width=.5\textwidth, height=.34\textwidth]{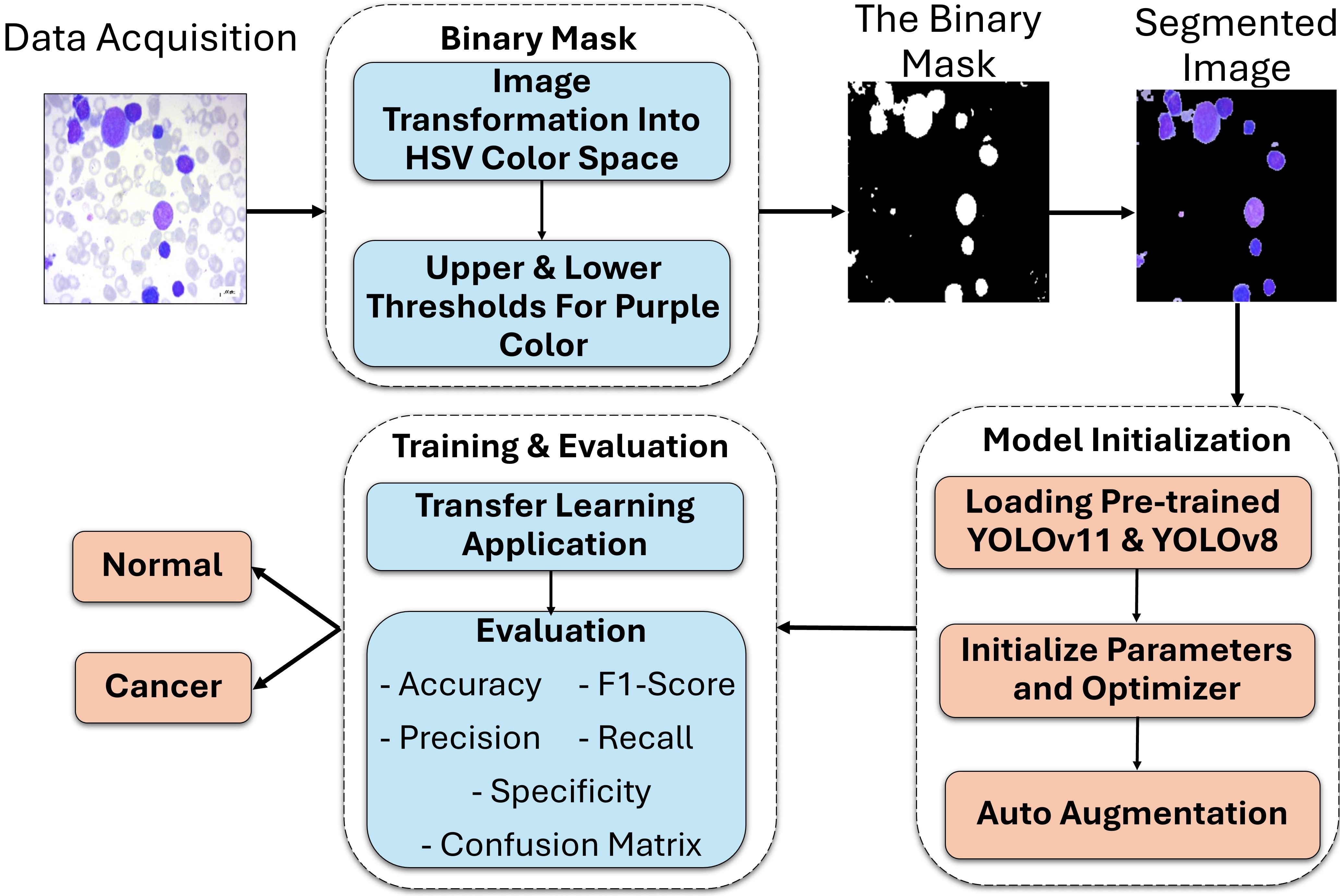}}
\caption{The implementation process including the data preparation and models training and evaluation}
\label{fig:workflow}
\end{figure}

The contributions of this research are summarized as follows:
\begin{enumerate}
\item Up to our knowledge, this is the first work to utilize YOLOv11 for ALL blood cancer detection.
        \item The integration of two datasets to improve generalization across different sample types.
            \item The classification of white blood cells as malignant or benign including hematogones is addressed.
                     \end{enumerate}
The structure of the paper is as follows. Section~\ref{sec:dataset} discusses the dataset used in this paper. Section \ref{sec:Implementation} goes through the methodologies and several deep learning models used. The performance metrics are discussed in section \ref{sec:metrics}, and the results of YOLOv11, YOLOv8  are found in section~\ref{sec:results}. Section~\ref{sec:relatedwork} discusses the related work done in this field, followed by our results  compared with other works in section~\ref{sec:comparison}. Finally, the conclusion is given in section~\ref{sec:conclusion}.

\section{Datasets and Data Collections}\label{sec:dataset}
Available datasets are divided into two types: single-cell and multi-cell datasets. Single-cell datasets typically contain images with a single white blood cell per image, whereas multi-cell datasets depict multiple cells within each sample. Since multi-cell datasets better represent real-life scenarios when working with blood cells, we chose to focus on them. The two datasets selected for this study are the Acute Lymphoblastic Leukemia (ALL) image dataset from Kaggle ~\cite{Ghaderzadeh2021} and ALL-IDB1 ~\cite{Genovese2023}, both of which contain multiple white blood cells per sample.

ALL image dataset, which contains 3,256 images in total, divided into four categories: Benign, Early, Pre, and Pro. The benign class includes Hematogones, a condition where lymphoid cells accumulate in a pattern similar to ALL but are non-cancerous and generally harmless. The dataset consists of 504 benign images and 2,752 malignant cells, further categorized into 985 early-stage samples, 963 pre-stage samples, and 804 pro-phase samples.

On the other hand, ALL-IDB1 dataset includes 108 images in total divided into 59 normal blood samples and 49 cancerous ones. This balance between normal and cancerous samples is crucial for the model to effectively learn the distinguishing features of ALL cells.

We decided to merge the residual normal cells from ALL-IDB1 with the benign cells from the ALL dataset into a single category called Normal. Similarly, we combined the Early, Pre, and Pro classes from the ALL dataset with the Cancer class from ALL-IDB1 into one category called Cancer. As a result, we focused on two classes: Normal and Cancer. This approach exposes the models to different datasets and various shapes of blast cells, allowing for more practical detection and classification.

\section{Models and Methodologies}\label{sec:Implementation}
The implementation is divided into several phases, as illustrated in Fig.~\ref{fig:workflow}. The first phase involves data preparation, where image segmentation techniques are applied to isolate the relevant elements. Next, the pretrained YOLOv11s and YOLOv8 models are loaded. These models enable data augmentation and experimentation with various optimizers and learning rates. In the final phase, the models are trained to fine-tune the pretrained weights for the specific task at hand.\\

\textbf{Dataset Preparation}
The dataset underwent preprocessing to enhance model performance by removing irrelevant elements like different backgrounds and unrelated blood components. Image segmentation was applied using OpenCV, converting images to the HSV color space and creating a binary mask to isolate white blood cells. To improve robustness and mitigate overfitting, various augmentation techniques were implemented. First, the augmentation parameter was set to true to allow the augmentation of the training data. Then mosaic augmentation applied which combines four different images into one during training. Setting it to 1.0 means that this augmentation is applied 100\% of the time. The degrees parameter that specifies the maximum degree of random rotation applied to images was set to 45 degrees. Additionally, horizontal flip was applied with 0.5 probability meaning that 50\% of the images will be flipped horizontally during training. Finally, the scale factor used to resize or zoom in/out on the image was given a value of 0.5, which means that images might be scaled by up to 50\%, either enlarging or shrinking them.

\textbf{Image Classification: }
The detection of blast cells can be approached in various ways, with image classification being one of the most common. Numerous deep learning architectures have been developed to support this task, with Convolutional Neural Networks (CNNs) being the most widely used for image and video datasets. Models like VGG , AlexNet, and GoogleNet (Inception) ~\cite{Szegedy2014} are all based on CNNs. In this paper, we focus on two versions of YOLO: YOLOv8 and YOLOv11.

\textbf{YOLOv8}: YOLOv8~\cite{Varghese2024} is a cutting-edge development in the YOLO object detection series~\cite{Redmon2016}, built upon the foundational CNN-based YOLO architecture. Its structure includes a backbone network, inspired by EfficientNet~\cite{Mehla2023}, for extracting multi-scale features from images, and a detection head based on NAS-FPN, which integrates these features for accurate object detection. YOLOv8 introduces several enhancements, such as the Focal Loss function, which prioritizes challenging examples during training, and Mixup, a data augmentation technique that blends images and labels for better generalization. Additionally, the model employs Average Precision Across Scales (APAS), a metric designed to assess detection accuracy across various object sizes, offering a more holistic evaluation compared to traditional metrics.

\textbf{YOLOv11}: YOLOv11~\cite{YOLOv11_2024}, the latest iteration in the YOLO series by Ultralytics, builds on its predecessors with significant improvements. It features an upgraded backbone and neck architecture for superior feature extraction, achieving higher detection accuracy while remaining computationally efficient with fewer parameters. The model is versatile, supporting tasks such as object detection, classification, segmentation, and pose estimation. Optimized for deployment on both edge devices and cloud platforms, YOLOv11 is designed to balance speed, accuracy, and adaptability across a wide range of applications.

\textbf{Methodology}:

First, transfer learning was applied using a pretrained YOLOv8 model, imported after installing the Ultralytics package, and then trained on our custom segmented dataset. The final model was based on 50 epochs of training, using the SGD optimizer with a learning rate of 0.001 and a batch size of 8. We experimented with more and less epochs.

Second, YOLOv11s, the latest version in the YOLO series, was also trained on our custom dataset. We experimented with the small version of the model to observe the performance on different versions of YOLO. This gave us the chance to understand the enhancements in the new version of the model. The best results were chosen based on training the model with 50 epochs, SGD optimizer, 0.001 learning rate, and 32 batch sizes.

Various tests were conducted using different optimizers and hyperparameters. Different optimizers, namely, AdamW, SGD, RMSProp, and Adam, were used to reach the optimal performance possible. We observed the accuracy, loss, and stability of the training to decide which optimizer works best. Moreover, different learning rates were tested to evaluate their effects and when the model converged better. Numerous epoch numbers were tested as well as diversity of batch sizes.

\section{Performance Metrics}\label{sec:metrics}
Accuracy is an overall indicator on how well the model performs taking into consideration the number of correctly identified samples out of all the given samples. This is represented by the summation of true positives and true negatives divided by the total number of examples consisting of True Positive(TP), True Negative(TN), False Positive (FP) and False Negative (FN) as expressed in Equation~\ref{eq:acc}:
\smallskip
\begin{equation}
 Accuracy = \frac{TP+TN}{TP+TN+FP+FN}
 \label{eq:acc}
\end{equation}
\smallskip
Presented in Equation~\ref{eq:prec}, the sample precision which is identified by the ratio of correctly classified instances to the total number of classified instances.
\smallskip
\begin{equation}
 Precision = \frac{TP}{TP+FP}
 \label{eq:prec}
\end{equation}
\smallskip
Recall, or Sensitivity, is calculated as the ratio of correctly identified instances to the total number of instances, as described in Equation~\ref{eq:recall}.
\begin{equation}
 Recall = \frac{TP}{TP+FN}
 \label{eq:recall}
\end{equation}
Another significant metric that contributed to our results is f1 score. It is obtained by calculating the harmonic mean of precision and recall, as illustrated in Equation~\ref{eq:f1}.
\begin{equation}
 F1 Score = 2*\frac{Precision * Recall}{Precision + Recall}
 \label{eq:f1}
\end{equation}

In addition to the previous indicators, we calculated the specificity using the formula in Equation~\ref{eq:spec}.It refers to the proportion of correctly identified negative instances among all actual negative cases. It reflects the model's ability to accurately classify instances from the opposite disease classes.
\begin{equation}
 Specificity = \frac{TN}{TN+FP}
 \label{eq:spec}
\end{equation}

\section{Experimental Results}\label{sec:results}
In this section, we evaluate the performance of our trained models using the metrics outlined earlier. We begin with YOLOv11s, which achieved 98.6\% accuracy on the validation dataset and 98.2\% on the test dataset. This is followed by YOLOv8, which attained a slightly lower testing accuracy of 98\%.

The experiments with YOLOv11 highlighted some key observations. The model trained with the SGD optimizer exhibited smoother training and validation curves, while the AdamW optimizer resulted in slightly higher accuracy. Additionally, increasing the batch size improved accuracy. However, extending the number of epochs beyond 50 led to a decline in accuracy, with 100 epochs or more proving counterproductive; therefore, we settled on 50 epochs as the optimal choice.

\begin{figure}[]
\centering
\begin{subfigure}[b]{0.6\columnwidth}
    \includegraphics[width=\linewidth]{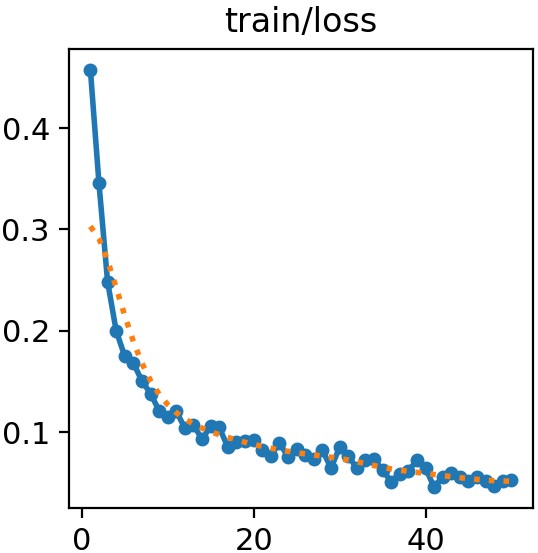}
    \caption{Train loss}
    \label{fig:train_loss_yolov11s_SGD}
\end{subfigure}%
\hfill
\begin{subfigure}[b]{0.6\columnwidth}
    \includegraphics[width=\linewidth]{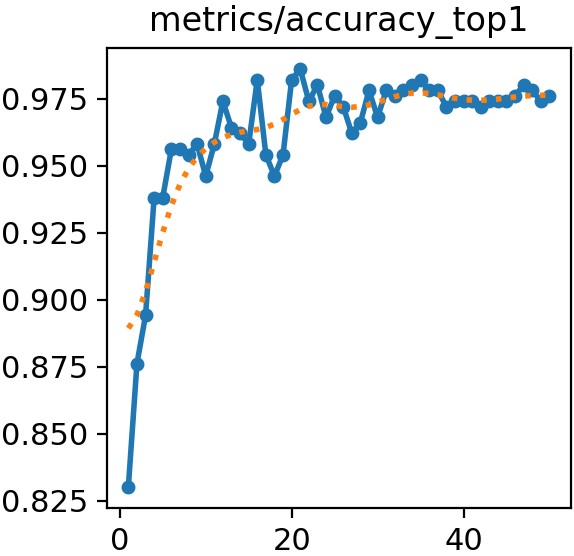}
    \caption{Accuracy (Top 1)}
    \label{fig:accuracy_yolov11s_SGD}
\end{subfigure}
\caption{Performance metrics for YOLOv11s using SGD optimizer. (a) Train loss, (b) Accuracy}
\label{fig:performance_metrics_yolov11s_SGD}
\end{figure}

\begin{figure}[t]
\centering
\begin{subfigure}[b]{0.7\columnwidth}
    \centering
    \includegraphics[width=5cm]{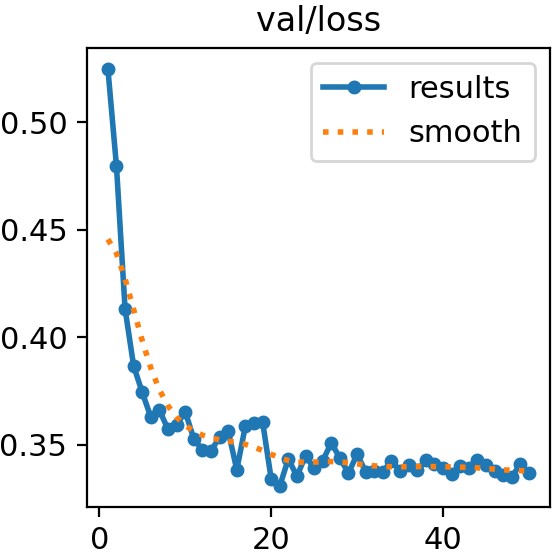}
    \caption{YOLOv11 validation loss}
    \label{fig:val_loss_yolov11s_SGD}
\end{subfigure}
\hfill

\begin{subfigure}[b]{0.7\columnwidth}
    \centering
    \includegraphics[width=5cm]{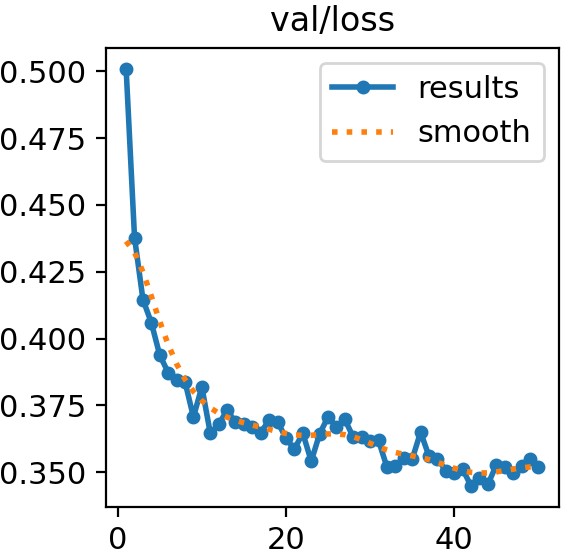}
    \caption{YOLOv8 validation loss}
    \label{fig:val_loss_yolov8s}
\end{subfigure}
\caption{The validation losses for both (a) YOLOv11 and (b) YOLOv8.}
\label{fig:validation_loss}

\end{figure}

The accuracy graph for the small version of YOLOv11 shown in Fig.~\ref{fig:accuracy_yolov11s_SGD} using SGD demonstrates improvement in the accuracy's progress with some fluctuations at the beginning. These variations decrease gradually as the number of epochs increases until the graph curve becomes more stable. It can be seen that the training and validation losses were declining steadily as the training advanced in Figures.~\ref{fig:train_loss_yolov11s_SGD} and ~\ref{fig:val_loss_yolov11s_SGD}.

\begin{figure}[]
\centering
\begin{subfigure}[b]{0.6\columnwidth}
    \includegraphics[width=\linewidth]{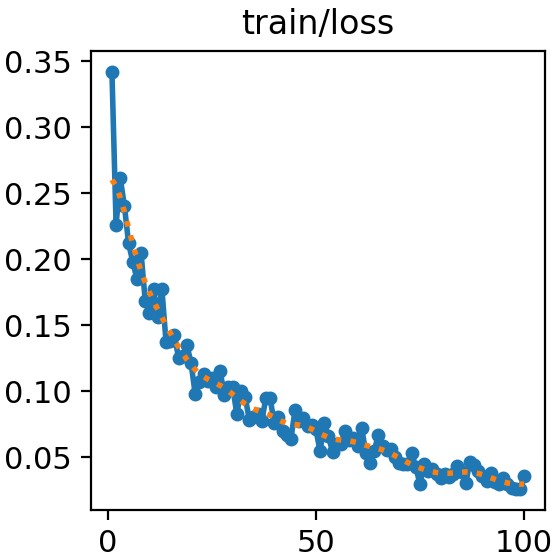}
    \caption{Train loss}
    \label{fig:train_loss_yolov11s}
\end{subfigure}%
\hfill
\begin{subfigure}[b]{0.6\columnwidth}
    \includegraphics[width=\linewidth]{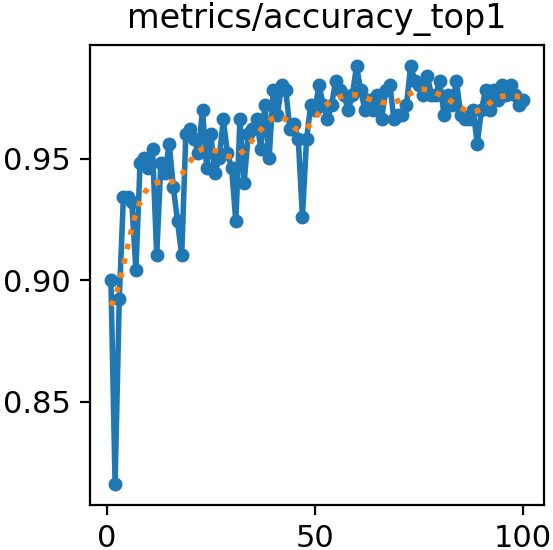}
    \caption{Accuracy (Top 1)}
    \label{fig:accuracy_yolov11s}
\end{subfigure}
\caption{Performance metrics for YOLOv11s using AdamW optimizer. (a) Train loss, (b) Accuracy}
\label{fig:performance_metrics_yolov11s}
\end{figure}

On the other hand, figure.~\ref{fig:accuracy_yolov11s} illustrates the performance of the model when trained with the AdamW optimizer. While it achieved a higher accuracy of 98.8\%, the training process exhibited significant fluctuations, indicating instability and potential overfitting. As a result, we selected the first model, trained with SGD, as the preferred choice due to its better generalization and more consistent performance.

\begin{figure}[htbp]
\centering
\includegraphics[ width=.42\textwidth,height=.25\textwidth]{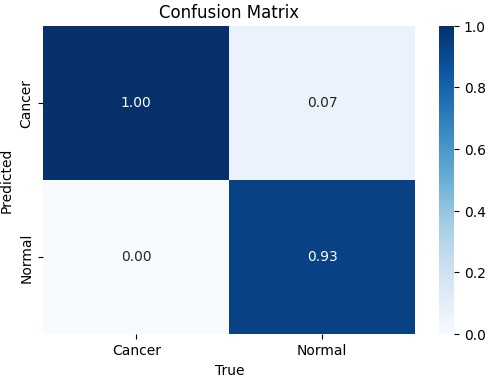}
\caption{Normalized confusion matrix for YOLOv11s.}
\label{fig:conf_matrix_yolov11s}
\end{figure}

The confusion matrix in Fig.~\ref{fig:conf_matrix_yolov11s} offers valuable insights into the YOLOv11s model's performance, highlighting which classes are accurately detected and where errors occur. This analysis helps identify areas for improvement to enhance the model's effectiveness. The matrix indicates that the model achieved high accuracy in detecting cancer across all stages, though it did misclassify 0.07\% of healthy white blood cells as cancerous. This misclassification is a result of the imbalanced data since the images of cancerous cells outnumber the healthy ones. Overall, the analysis validates the model's strong performance while pinpointing specific issues that require optimization.

\begin{figure}[]
\centering
\begin{subfigure}[b]{0.6\columnwidth}
    \includegraphics[width=\linewidth]{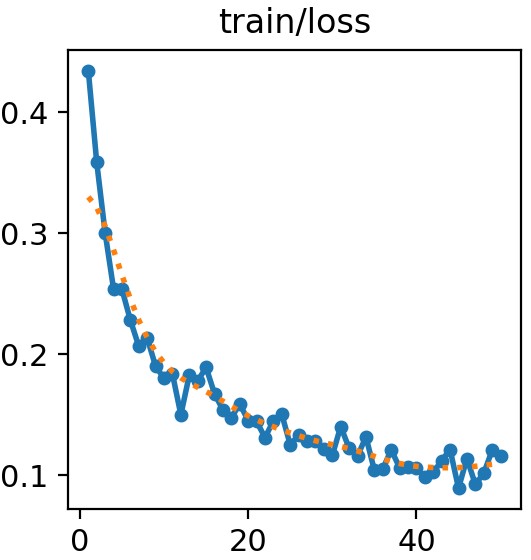}
    \caption{Train loss}
    \label{fig:train_loss_yolov8s}
\end{subfigure}%
\hfill
\begin{subfigure}[b]{0.6\columnwidth}
    \includegraphics[width=\linewidth]{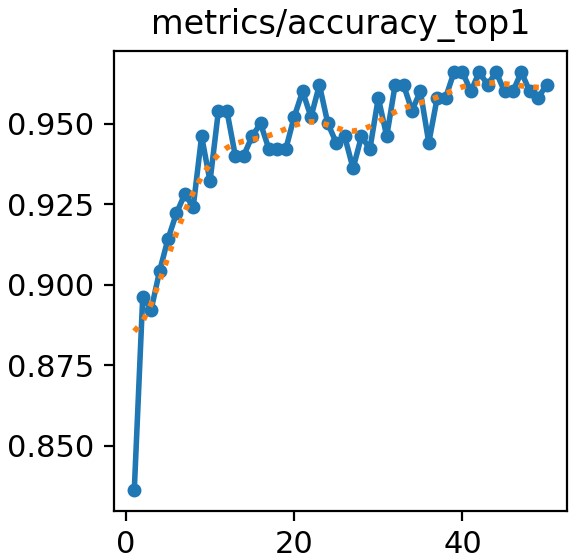}
    \caption{Accuracy (Top 1)}
    \label{fig:accuracy_yolov8s}
\end{subfigure}
\caption{Performance metrics for YOLOv8s using SGD optimizer. (a) Train loss, (b) Accuracy}
\end{figure}

Next, we examine the results of YOLOv8, illustrated in Figures ~\ref{fig:train_loss_yolov8s}, ~\ref{fig:accuracy_yolov8s} and ~\ref{fig:val_loss_yolov8s}. The model achieved an accuracy of 96.6\% on the validation dataset, which improved to 98\% when tested on the testing dataset. Compared to YOLOv11, the small version of YOLOv8 achieved slightly lower accuracy. The accuracy, training loss, and validation loss graphs for YOLOv8 exhibit patterns similar to those of YOLOv11, although YOLOv11 demonstrated more stable curves overall.

When comparing optimizers, YOLOv8 showed greater stability with SGD compared to AdamW. In terms of batch size, we tested values of 8, 16, 32, and 64, and observed that the model performed better with smaller batch sizes. As a result, we selected a batch size of 8 for our final model.

\begin{figure}[htbp]
\centering
\includegraphics[width=.42\textwidth,height=.27\textwidth]{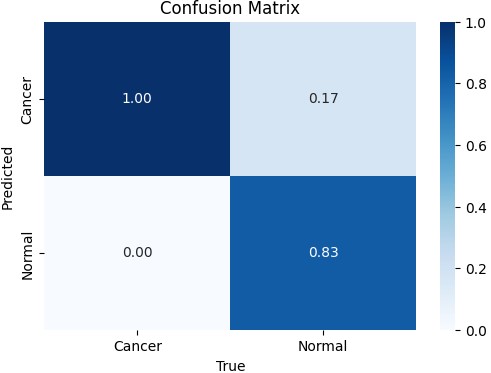}
\caption{Confusion matrix for YOLOv8s.}
\label{fig:conf_matrix_yolov8s}
\end{figure}

The confusion matrix for YOLOv8, shown in Figure~\ref{fig:conf_matrix_yolov8s}, highlights the reason for the model's slightly lower accuracy. It reveals a higher number of misclassifications, particularly in the normal cell samples, with a misclassification rate of 0.17\%.

Table~\ref{tab:performance_metrics} highlights the key findings of this study by evaluating the performance of YOLOv11 and YOLOv8 through different metrics. It is notable that YOLOv11's accuracy was 0.2\% higher than that of YOLOv8, which is significant when working with sensitive tasks like cancer detection. The rest of the metrics' values prove that the 11th version outperforms the 8th one. The specificity values show that YOLOv11 was 10.4\% better at recognizing true negatives representing cases of normal cells.

\section{Related Work}\label{sec:relatedwork}
\medskip

Using a deep CNN, Hosseini et al.~\cite{Hosseini2023} attempted to identify cases of B-cell acute lymphoblastic leukemia (B-ALL), including its subtypes. He compared the effectiveness of three lightweight CNN models (EfficientNetB0, MobileNetV2, and NASNet Mobile) utilizing the training and testing data after employing K-means clustering and segmentation for image preprocessing on a dataset of benign and malignant B-ALL patients. The accuracy of the models was eventually improved by combining segmented and original images and feeding them as inputs through two channels to extract the maximal feature space. MobileNetV2 was chosen because to its 100\% accuracy and minimal size, which makes it appropriate for use on mobile devices.

Talaat et al.~\cite{Talaat2023} proposed the A2M-LEUK algorithm for leukemia cell classification, incorporating image preprocessing, CNN feature extraction, and an attention mechanism-based machine learning approach. The model employed classifiers such as SVM and neural networks and was evaluated on the C-NMC 2019~\cite{Mourya2019} dataset. It outperformed other methods like KNN, SVM, Random Forest, and Naïve Bayes in precision, recall, accuracy, and specificity, achieving over 100\% in all metrics. However, the specific classification model used with A2M-LEUK was not detailed in the paper.

\begin{table}[htbp]
\centering
\caption{Performance metrics for different models}
\begin{tabular}{|l|c|c|c|c|c|}
\hline
Model & Accuracy & F1 & Precision & Recall & Specificity \\
\hline
YOLOv11s & 98.2 & 99.2 & 98.6 & 99.8 & 93 \\
YOLOv8s & 98 & 98 & 96.5 & 99.6 & 82.6 \\
\hline
\end{tabular}
\label{tab:performance_metrics}
\end{table}
\medskip

Sampathila et al.~\cite{Sampathila2022} built a new CNN model called ALLNET which was trained on the C-NMC 2019 dataset. The model differentiated between healthy and blast cells, and the study exploited image segmentation and data augmentation preprocessing methodologies. The CNN comprised convolution, pooling, batch normalization, dropout and fully connected layers. The maximum accuracy reached was 95.54\%.

Yan ~\cite{Yan2024} worked with the single-cell dataset CNMC-2019 ~\cite{Mourya2019} to classify normal and cancerous white blood cells using three models: YOLOv4, YOLOv8, and a CNN. Data augmentation was applied to the CNN and YOLOv4 models. The CNN model, featuring convolutional layers, max-pooling layers, and ReLU activation, achieved 93\% accuracy, while YOLOv4 and YOLOv8 surpassed 95\%.

Devi et al. ~\cite{Devi2024} combined custom-designed and pretrained CNN architectures to detect ALL in the augmented ALL image dataset ~\cite{Ghaderzadeh2021}. The custom CNN extracted hierarchical features, while VGG-19 extracted high-level features and performed classification, achieving 97.85\% accuracy. In contrast, Khosrosereshki ~\cite{Khosrosereshki2017} used image processing and a Fuzzy Rule-Based inference system for this task.

Rahmani et al. ~\cite{Rahmani2024} utilized the C-NMC 2019 dataset, applying preprocessing techniques such as grayscaling and masking, followed by feature extraction via transfer learning using VGG19, ResNet50, ResNet101, ResNet152, EfficientNetB3, DenseNet-121, and DenseNet-201. Feature selection employed Random Forest, Genetic Algorithms, and Binary Ant Colony Optimization. The classification, done through a multilayer perceptron, achieved slightly above 90\% accuracy.

Kumar et al. ~\cite{Kumar2020} focused on classifying different blood cancers, including ALL and Multiple Myeloma, in white blood cells. After preprocessing and augmentation, feature selection was done using SelectKBest. Their model comprised two blocks with convolutional and max-pooling layers, followed by fully connected and classification layers, achieving 97.2\% accuracy.

Saikia et al. ~\cite{Saikia2024} introduced VCaps-Net, a fine-tuned VGG16 combined with a capsule network for ALL detection. Using the ALL-IDB1 dataset ~\cite{Genovese2023} and a private dataset, VCaps-Net maintained spatial relationships in images through capsule vectors, avoiding the loss often caused by MaxPooling, and achieved 98.64\% accuracy.

\medskip

\section{Comparison Study}\label{sec:comparison}

\medskip
The performance of YOLOv11 proved to be slightly better than that of YOLOv8, achieving better accuracy, which can be significant in cancer diagnosis. Table~\ref{tab:ALL_detection_comparison} demonstrates a comparison between our findings and some of the previous studies. As indicated in Table~\ref{tab:ALL_detection_comparison}, we compare our image classification techniques for Acute Lymphoblastic Leukemia (ALL) with current methods in this section. The table shows that Yan's YOLOv8 model achieved 96\% accuracy on the C-NMC 2019 dataset, whereas our method using YOLOv11s achieved 98.2\% accuracy, while our YOLOv8s model achieved 98\%. In addition, both of our models surpassed Sampathila's, Devi's and Kumar's CNNs. Although Vcaps-Net outperforms our chosen models, YOLOv11 utilizing the AdamW optimizer achieved a better accuracy of 98.8\%.

\begin{table}[ht]
    \centering
    \caption{Comparison of Different Approaches for Detecting Acute Lymphoblastic Leukemia (ALL)}
   \begin{tabular}{ |p{1.75cm}|p{2cm}|p{1.8cm}|p{1.6cm}|}
        \hline
        \hline
        &&&\\
        \textbf{Study} & \textbf{Methodology} & \textbf{Accuracy} & \textbf{Dataset}\\
        &&&\\
        \hline
        \hline
        Sampathila et al.~\cite{Sampathila2022}& ALLNET &95.54\% & C-NMC 2019\\
        \hline
        Yan~\cite{Yan2024}&YOLOv4&YOLOv4: 98\% &\\
        &YOLOv8&YOLOv8: 96\% &C-NMC 2019\\
        &CNN &CNN: 92\%&\\
        \hline
        Devi et al.~\cite{Devi2024}& Custom + pretrained CNN & 97.85\% & ALL dataset \\
        \hline
        Saikia et al.~\cite{Saikia2024}& VCaps-Net & 98.64\% & ALL-IDB1 \\
        \hline
         Kumar et al.~\cite{Kumar2020}& Custom CNN  & 97.2\% & Custom dataset \\
        \hline
        Our study&YOLOv11s&YOLOv11s: 98.2\%&ALL-IDB1\\
        &YOLOv8s&YOLOv8s: 98\%&+ ALL dataset\\
        \hline

    \end{tabular}
    \label{tab:ALL_detection_comparison}
\end{table}
\section{Conclusion}\label{sec:conclusion}
In conclusion, the integration of AI in the medical field is a massive step in the advancement of the health system and services provided to patients. This study was able to detect the presence of ALL in blood even at early stages using YOLOv11 and YOLOv8.

\bibliographystyle{IEEEtran}
\bibliography{conf_bloodcancerV13.bbl}

\begin{thebibliography}{10}
\providecommand{\url}[1]{#1}
\csname url@samestyle\endcsname
\providecommand{\newblock}{\relax}
\providecommand{\bibinfo}[2]{#2}
\providecommand{\BIBentrySTDinterwordspacing}{\spaceskip=0pt\relax}
\providecommand{\BIBentryALTinterwordstretchfactor}{4}
\providecommand{\BIBentryALTinterwordspacing}{\spaceskip=\fontdimen2\font plus
\BIBentryALTinterwordstretchfactor\fontdimen3\font minus
  \fontdimen4\font\relax}
\providecommand{\BIBforeignlanguage}[2]{{%
\expandafter\ifx\csname l@#1\endcsname\relax
\typeout{** WARNING: IEEEtran.bst: No hyphenation pattern has been}%
\typeout{** loaded for the language `#1'. Using the pattern for}%
\typeout{** the default language instead.}%
\else
\language=\csname l@#1\endcsname
\fi
#2}}
\providecommand{\BIBdecl}{\relax}
\BIBdecl

\bibitem{Sandler2019}
\BIBentryALTinterwordspacing
M.~Sandler, A.~Howard, M.~Zhu, A.~Zhmoginov, and L.-C. Chen, ``Mobilenetv2:
  Inverted residuals and linear bottlenecks,'' 2019. [Online]. Available:
  \url{https://arxiv.org/abs/1801.04381}
\BIBentrySTDinterwordspacing

\bibitem{Vaswani2023}
\BIBentryALTinterwordspacing
A.~Vaswani, N.~Shazeer, N.~Parmar, J.~Uszkoreit, L.~Jones, A.~N. Gomez,
  L.~Kaiser, and I.~Polosukhin, ``Attention is all you need,'' 2023. [Online].
  Available: \url{https://arxiv.org/abs/1706.03762}
\BIBentrySTDinterwordspacing

\bibitem{Redmon2016}
J.~Redmon, S.~Divvala, R.~Girshick, and A.~Farhadi, ``You only look once:
  Unified, real-time object detection,'' in \emph{2016 IEEE Conference on
  Computer Vision and Pattern Recognition (CVPR)}, 2016.

\bibitem{Ghaderzadeh2021}
M.~Ghaderzadeh, M.~Aria, A.~Hosseini, F.~Asadi, D.~Bashash, and H.~Abolghasemi,
  ``A fast and efficient cnn model for b‐all diagnosis and its subtypes
  classification using peripheral blood smear images,'' \emph{International
  Journal of Intelligent Systems}, Nov 2021.

\bibitem{Mourya2019}
S.~Mourya, S.~Kant, P.~Kumar, A.~Gupta, and R.~Gupta, ``All challenge dataset
  of isbi 2019 (c-nmc 2019) (version 1),'' The Cancer Imaging Archive, 2019.

\bibitem{who2023}
A.~S. of~Hematology, ``Leukemia,''
  https://www.hematology.org/education/patients/blood-cancers/leukemia, 2023,
  accessed: 2024-09-25.

\bibitem{YOLOv11_2024}
\BIBentryALTinterwordspacing
Ultralytics, ``Yolov11 - key features,'' 2024, accessed: October 8, 2024.
  [Online]. Available:
  \url{https://docs.ultralytics.com/models/yolo11/#key-features}
\BIBentrySTDinterwordspacing

\bibitem{Varghese2024}
R.~Varghese and S.~M., ``Yolov8: A novel object detection algorithm with
  enhanced performance and robustness,'' in \emph{2024 International Conference
  on Advances in Data Engineering and Intelligent Computing Systems (ADICS)},
  Chennai, India, 2024, pp. 1--6.

\bibitem{Genovese2023}
A.~Genovese, V.~Piuri, K.~N. Plataniotis, and F.~Scotti, ``{DL4ALL: Multi-Task
  Cross-Dataset Transfer Learning for Acute Lymphoblastic Leukemia
  Detection},'' \emph{IEEE Access}, vol.~11, pp. 65\,222--65\,237, 2023.

\bibitem{Szegedy2014}
\BIBentryALTinterwordspacing
C.~Szegedy, W.~Liu, Y.~Jia, P.~Sermanet, S.~Reed, D.~Anguelov, D.~Erhan,
  V.~Vanhoucke, and A.~Rabinovich, ``Going deeper with convolutions,'' 2014.
  [Online]. Available: \url{https://arxiv.org/abs/1409.4842}
\BIBentrySTDinterwordspacing

\bibitem{Mehla2023}
N.~Mehla, Ishita, R.~Talukdar, and D.~K. Sharma, ``Object detection in
  autonomous maritime vehicles: Comparison between yolo v8 and efficientdet,''
  in \emph{International Conference on Data Science and Network
  Engineering}.\hskip 1em plus 0.5em minus 0.4em\relax Singapore: Springer
  Nature Singapore, 2023, pp. 125--141.

\bibitem{Hosseini2023}
A.~Hosseini \emph{et~al.}, ``A mobile application based on efficient
  lightweight cnn model for classification of b-all cancer from non-cancerous
  cells: A design and implementation study,'' \emph{Informatics in Medicine
  Unlocked}, vol.~39, pp. 101\,244--101\,244, Jan 2023.

\bibitem{Talaat2023}
F.~M. Talaat and S.~A. Gamel, ``A2m-leuk: attention-augmented algorithm for
  blood cancer detection in children,'' \emph{Neural Computing and
  Applications}, vol.~35, no.~24, pp. 18\,059--18\,071, Jun 2023.

\bibitem{Sampathila2022}
N.~Sampathila, K.~Chadaga, N.~Goswami, R.~Chadaga, M.~Pandya, S.~Prabhu,
  M.~Bairy, S.~Katta, D.~Bhat, and S.~Upadya, ``Customized deep learning
  classifier for detection of acute lymphoblastic leukemia using blood smear
  images,'' \emph{Healthcare (Basel)}, vol.~10, no.~10, p. 1812, Sep 2022.

\bibitem{Yan2024}
E.~Yan, ``Detection of acute myeloid leukemia using deep learning models based
  systems,'' in \emph{IFMBE Proceedings}, Jan 2024, pp. 421--431.

\bibitem{Devi2024}
J.~R. Devi, P.~S. Kadiyala, S.~Lavu, N.~Kasturi, and L.~Kosuri, ``Enhancing
  acute lymphoblastic leukemia classification with a rapid and effective cnn
  model,'' in \emph{2024 Third International Conference on Distributed
  Computing and Electrical Circuits and Electronics (ICDCECE)}, Ballari, India,
  2024, pp. 1--6.

\bibitem{Khosrosereshki2017}
M.~A. Khosrosereshki and M.~B. Menhaj, ``A fuzzy based classifier for diagnosis
  of acute lymphoblastic leukemia using blood smear image processing,'' in
  \emph{2017 5th Iranian Joint Congress on Fuzzy and Intelligent Systems
  (CFIS)}, Qazvin, Iran, 2017, pp. 13--18.

\bibitem{Rahmani2024}
\BIBentryALTinterwordspacing
A.~M. Rahmani \emph{et~al.}, ``A diagnostic model for acute lymphoblastic
  leukemia using metaheuristics and deep learning methods,'' arXiv.org, 2024,
  accessed: Sep. 25, 2024. [Online]. Available:
  \url{https://arxiv.org/abs/2406.18568}
\BIBentrySTDinterwordspacing

\bibitem{Kumar2020}
D.~Kumar, N.~Jain, A.~Khurana, S.~Mittal, S.~C. Satapathy, R.~Senkerik, and
  J.~D. Hemanth, ``Automatic detection of white blood cancer from bone marrow
  microscopic images using convolutional neural networks,'' \emph{IEEE Access},
  vol.~8, pp. 142\,521--142\,531, 2020.

\bibitem{Saikia2024}
R.~Saikia, A.~Sarma, K.~M. Singh, and S.~S. Devi, ``Vcaps-net: Fine-tuned vgg16
  with capsule network for acute lymphoblastic leukemia detection on a diverse
  dataset,'' in \emph{2024 6th International Conference on Energy, Power and
  Environment (ICEPE)}, 2024, pp. 1--6.

\end{thebibliography}

\end{document}